\documentclass[
	usletter
	]{IEEEtran}

\usepackage{cite}
\usepackage{textcomp}
\def\BibTeX{{\rm B\kern-.05em{\sc i\kern-.025em b}\kern-.08em
		T\kern-.1667em\lower.7ex\hbox{E}\kern-.125emX}}

\usepackage[utf8]{inputenc} %
\usepackage[T1]{fontenc}    %
\usepackage{hyperref}       %
\usepackage{url}            %
\usepackage{booktabs}       %
\usepackage{amsfonts}       %
\usepackage{nicefrac}       %
\usepackage{microtype}      %
\usepackage{algorithm}
\usepackage{algorithmic}
\usepackage{microtype}
\usepackage{graphicx}
\usepackage{subcaption}
\usepackage{xcolor}
\usepackage{float}
\usepackage{amsmath}
\usepackage{enumitem}
\usepackage{amsmath}
\usepackage{amssymb}
\usepackage{amsthm}
\usepackage[capitalize,noabbrev]{cleveref}
\usepackage[bottom]{footmisc}

\usepackage{mathtools}

\DeclareMathOperator*{\argmax}{arg\,max}

\newlength\tindent
\newcommand{\R}{\mathbb{R}}

\newcommand{\E}{\mathbb{E}}

\renewcommand{\L}{\mathcal{L}}

\newcommand{\I}{\mathcal{I}}
\renewcommand{\P}{\mathcal{P}}
\newcommand{\prox}{\text{prox}}
\newcommand{\xh}{\hat{x}}

\begin{document}
	
\title{Gradient-Based Learning of Discrete Structured Measurement Operators for Signal Recovery}

\author{Jonathan Sauder, Martin Genzel, and Peter Jung
	\thanks{This work was presented in part at the NeurIPS 2021 Workshop ``Deep Learning and Inverse Problems'' \cite{sauder2021learning}. M.~G.~acknowledges support by the DFG Priority Programme DFG-SPP 1798 Grant DI 2120/1-1.
		P.~J.~was partially funded by the German Federal Ministry of Education and Research (BMBF) in the framework of the international future AI lab ``AI4EO -- Artificial Intelligence for Earth Observation: Reasoning, Uncertainties, Ethics and Beyond'' (Grant number: 01DD20001).}
	\thanks{J.~S.~is with the ECEO Laboratory, EPFL, CH-1951 Sion, Switzerland (e-mail: jonathan.sauder@epfl.ch).}
	\thanks{M.~G.~is with the Helmholtz-Zentrum Berlin, 14109 Berlin, Germany (e-mail: martin.genzel@helmholtz-berlin.de)}
	\thanks{P.~J.~is with the CommIT Laboratory, Technische Universität Berlin, 10587 Berlin, Germany (e-mail: peter.jung@tu-berlin.de)}
	\thanks{The current version of the manuscript has been accepted to IEEE Journal on Selected Areas in Information Theory, Digital Object Identifier 10.1109/JSAIT.2022.3221644}
	\thanks{\textcopyright2022 IEEE.  Personal use of this material is permitted.  Permission from IEEE must be obtained for all other uses, in any current or future media, including reprinting/republishing this material for advertising or promotional purposes, creating new collective works, for resale or redistribution to servers or lists, or reuse of any copyrighted component of this work in other works.}
	\vspace{-\baselineskip}}

\maketitle

\begin{abstract}
Countless signal processing applications include the reconstruction of signals from few indirect linear measurements. 
The design of effective measurement operators is typically constrained by the underlying hardware and physics, posing a challenging and often even discrete optimization task.
While the potential of gradient-based learning via the unrolling of iterative recovery algorithms has been demonstrated, it has remained unclear how to leverage this technique when the set of admissible measurement operators is structured and discrete.
We tackle this problem by combining unrolled optimization with Gumbel reparametrizations, which enable the computation of low-variance gradient estimates of categorical random variables. 
Our approach is formalized by GLODISMO (Gradient-based Learning of \mbox{DIscrete} Structured Measurement Operators). 
This novel method is easy-to-implement, computationally efficient, and extendable due to its compatibility with automatic differentiation. 
We empirically demonstrate the performance and flexibility of GLODISMO in several prototypical signal recovery applications, verifying that the learned measurement matrices outperform conventional designs based on randomization as well as discrete optimization baselines.
\end{abstract}


\section{Introduction}
Linear measurement operators adhering to a specific structure due to physical constraints of the observation process or the hardware of the measuring device are ubiquitous in signal processing. Corresponding structural constraints appear in numerous relevant applications, like magnetic resonance imaging with masked Fourier matrices \cite{csmri}, communication and remote sensing tasks \cite{zhu2010tomographic}, single-pixel imaging \cite{singlepixel}, compressed sensing with expander-graphs \cite{foucart2013invitation}, or pooling matrices for group testing \cite{petercovid19}. The optimal design of such measurement operators---which typically reside in a discrete subset---to improve the performance of downstream tasks poses great computational challenges.  While it is often easy to create a suitable random mask, it is not obvious how to optimize the measurement matrix in a way that is both efficient and respects the feasibility constraints. Classical approaches commonly use discrete optimization to find such sets, as no gradients can be directly computed.

On the other hand, gradient-based optimization via backpropagation through massive nonlinear computational graphs has shown impressive results in the field of machine learning. 
In order to enable these techniques even in non-differentiable settings, recent work has developed a variety of approaches for computing low-variance estimates of gradients. In this context, a promising concept is given by Gumbel reparametrizations \cite{gumbel, concrete}. These allow for estimating the gradients of categorical random variables by simply adding component-wise independent and identically distributed (i.i.d.) noise from a Gumbel distribution to the logits. This easy-to-implement method has proven successful in many machine learning applications, including graph learning \cite{gggan}, discrete representation learning \cite{learned_discrete1}, and neural architecture search \cite{nas}. However, Gumbel reparametrizations have found little attention in the signal processing and inverse problems communities so far, except for magnetic resonance imaging \cite{huijben2019deep, huijben2020learning, bahadir2019adaptive, bahadir2020deep}. The present work aims at using Gumbel reparametrizations to fuse gradient-based learning with the design of structured measurement operators that are constrained to a discrete set.

The actual signal recovery problem is typically solved by convex optimization methods.
It is well-established that the computational graph of common iterative optimization schemes can be unrolled to obtain a neural network that can be readily backpropagated through \cite{lista,Monga:2021}.
This strategy allows the solver to adapt to (training) data and has proven very powerful, especially in reducing the number of iterations for convex programs by orders of magnitudes \cite{lista, alista, na-alista}. 
Importantly, unrolling also enables the computation of gradients with respect to the measurement operator and other parameters of the reconstruction algorithm. However, it has remained unclear how this technique can be leveraged when the underlying parameter set is structured and discrete. This important shortcoming has made unrolled optimization infeasible for the design of practicable measurement operators. In this paper, we present a novel approach to tackle this problem.
Our main contributions can be summarized as follows:
\begin{itemize}
\item 
	We propose an efficient and easy-to-implement method for Gradient-based Learning of \mbox{DIscrete} Structured Measurement Operators (GLODISMO), which combines the concepts of unrolled optimization and Gumbel reparametrizations.
\item 
	We successfully apply GLODISMO to several prototypical signal recovery tasks, namely single-pixel imaging, compressed sensing with left-$d$-regular graphs, and pooling matrices for group testing.
	In particular, conventional baselines relying on randomization and discrete optimization are significantly outperformed.
\item
	GLODISMO provides a flexible learning framework, rather than a rigid algorithm.
	For example, we demonstrate that our basic approach (Algorithm~\ref{alg:algorithm}) can be easily extended to produce measurement operators that enable fast transforms.
	Moreover, the compatibility with automatic differentiation unlocks many potential applications beyond signal recovery.
\end{itemize}

\section{Background \& Related Work}

\subsection{Linear Inverse Problems \& Compressed Sensing}

In linear inverse problems, the basic task is to recover an unknown target signal $x\in \R^n$ from indirect observations of the form $y=\Phi x$ (mostly also contaminated by additive noise), where $\Phi\in \R^{m \times n}$ is a known measurement matrix. The number of measurements $m$ is usually much smaller than the ambient dimension of the signal $n$, making the inverse problem ill-posed and only solvable under prior knowledge about the underlying signals.

A prominent class of such inverse problems is given by compressed sensing, in which the signal is assumed to be sparse. The seminal papers in this field \cite{candes2006stable, donoho2006compressed} have shown that robust and stable recovery can be achieved with computationally tractable algorithms if the measurement matrix fulfills appropriate conditions.
For example, when $x$ is $s$-sparse, i.e., $x \in \Sigma_s^n := \{x \in\nobreak \R^n \, | \, \|x\|_0 \leq\nobreak s\}$, unique reconstruction is possible via convex optimization, given that $\Phi$ fulfills the null space property \cite{foucart2013invitation}. The convex program that is to be solved corresponds to a LASSO problem \cite{lasso} with hyperparameter $\lambda$:
\begin{equation}\label{lasso}
	\min_{\hat{x}} \ \|\Phi \hat{x} - y\|_2^2 + \lambda\|\hat{x}\|_1.
\end{equation}
When the signal $x$ is not sparse by itself but with respect to a basis transform $\Psi \in \R^{n \times n}$, the reparametrization $\overline{x} = \Psi \hat{x}$ can be incorporated into the reconstruction algorithm, so that the problem \eqref{lasso} yields the synthesis formulation:
\begin{equation*}
	\min_{\overline{x}} \ \|\Phi \Psi^* \overline{x} - y\|_2^2 + \lambda\|\overline{x}\|_1.
\end{equation*}
For natural images, useful sparsifying transforms include wavelets \cite{haar, wavelet, daubechies}, shearlets \cite{shearlet}, discrete cosine transforms \cite{dct}, and total variation synthesis \cite{tv}.

As mentioned above, the success of compressed sensing particularly relies on desirable properties of the measurement matrix $\Phi$ (or~$\Phi \Psi^*$), such as the null space property. On the theoretical side, it has been shown that various random matrices with $m \in O(s \log(n/s))$ satisfy such criteria with high probability. Prototypical examples are Gaussian matrices with $\Phi_{ij} \sim \mathcal{N}(0, \frac{1}{m})$ or Bernoulli matrices whose entries take values in $\{\pm \tfrac{1}{\sqrt{m}}\}$ with equal probability \cite{foucart2013invitation}. 
Similarly, random i.i.d.~(scaled) binary matrices fulfill the null space property in this regime with overwhelmingly high probability~\cite{kueng:16:nnls}.
Guarantees are also available for randomly subsampled Fourier matrices \cite{rv08}, which is accompanied by highly efficient matrix-vector multiplications due to the fast Fourier transform.
While the randomization of measurement operators is key to the theoretical analysis of sparse recovery problems, this technique is neither necessary nor practicable in most real-world applications.

A widely-used class of algorithms for solving LASSO-type problems are gradient-based methods. 
For example, an iterative proximal scheme of the following type can be used to solve \eqref{lasso}:
\begin{equation}\label{proximal}
	\xh^{(t+1)} = \prox_{\lambda\| \cdot \|_1}\Big(\xh^{(t)} - \gamma \nabla_{\xh^{(t)}}\big(\|y - \Phi \xh^{(t)}\|_2^2 \big) \Big),
\end{equation}
where $t = 0, 1, \dots, T-1$ and the proximal operator of the \mbox{$\lambda$-weighted} $\ell_1$-norm corresponds to element-wise soft thresholding:
\begin{equation}\label{soft_thresholding}
	\prox_{\lambda\| \cdot \|_1}(v) := 
	\text{sign}(v) \cdot \max \{|v|-\lambda, 0\}, \quad v \in \R.
\end{equation}
The algorithm described in \eqref{proximal} is known as the Iterative Soft Thresholding Algorithm (ISTA) \cite{ista}. 
Another popular method is based on Iterative Hard Thresholding (IHT) \cite{iht}, in which the proximal operator in \eqref{soft_thresholding} is replaced by a projection onto the set of $s$-sparse vectors $\Sigma_s^n$. This corresponds to a hard-thresholding operator, which sets all entries of a vector to zero except for those with the $s$ largest absolute values.

\subsection{Unrolled Optimization in Linear Inverse Problems}

It is well-established that many iterative optimization schemes can be viewed as a neural network via unrolling the computational graph of a finite number of iterations~$T$. Once unrolled, the tunable parameters $\theta$ of a given reconstruction algorithm\footnote{For better readability, we do not indicate the explicit dependency on $\Phi$, but always understand $f_\theta=f_{\theta,\Phi}$.} 
$f_\theta:\R^m \rightarrow \R^n$ can be fit to a (training) dataset by minimizing a loss function \mbox{$\L : \R^n \times \R^n \to \R$} via (stochastic) gradient-based optimization with observations perturbed by random noise $z$: 

\begin{equation}\label{unrolling}
	\min_{\theta} \ \E_{x, z} \Big[\L(f_\theta(\Phi x + z), x)\Big].
\end{equation}
In the context of linear inverse problems, a prominent example of unrolled optimization is Learned ISTA (LISTA) \cite{lista}, where ISTA is unrolled and the involved matrices, step-sizes, and thresholds are learned in an end-to-end fashion via \eqref{unrolling}. Such a data-driven approach can reduce the number of iterations required for convergence by orders of magnitude. LISTA has inspired a line of research on unrolled optimization algorithms for compressed sensing. In Analytic LISTA (ALISTA), for example, the involved matrices are computed analytically and remain fixed, while only the step-sizes and thresholds are learned \cite{alista}. This significantly reduces the number of trainable parameters, and theoretical guarantees known for ISTA are retained, but the performance of a fully learned LISTA is still matched.
Going one step further, Neurally Augmented \mbox{ALISTA} (\mbox{NA-ALISTA}) equips \mbox{ALISTA} with a recurrent neural network that enables adaptive step-sizes and thresholds during reconstruction, depending on the target vector \cite{na-alista}.

While compressed sensing theory focuses on recovery guarantees for generic sparse vectors, real-world signals often carry additional structure, for which random measurement matrices may not be optimal. This shortcoming has inspired the search for data-driven optimization of measurement operators. Unrolled optimization provides a natural approach in this regard, since $\Phi$ can be directly included in the above end-to-end training procedure, i.e., the objective in \eqref{unrolling} is also minimized over $\Phi$. This is equivalent to learning an autoencoder neural network in which the encoder (a single linear layer) becomes the measurement matrix. In the context of compressed sensing, this has been explored by Wu et al.~\cite{steps}, where a measurement matrix (initialized with Gaussian entries) is learned by backpropagating through a fixed number of steps of subgradient descent. This scheme has been also investigated in a compressive learning setting by Adler et al.~\cite{eladcompressive}, where a neural network classifier is learned with a linear first layer. Both works have demonstrated that end-to-end learning can provide decent measurement matrices when there are no additional constraints on the involved operator. Similarly, LISTA and its variants use end-to-end learning on unconstrained tuning parameter sets, initialized via Gaussian random variables. However, in real-world applications, the measurement matrix often must follow specific structures imposed by the underlying hardware and physics of the problem. In magnetic resonance imaging, for example, the measurement operator is a subsampled Fourier matrix \cite{csmri}, whereas in single-pixel imaging \cite{singlepixel}, only binary matrices are admissible. In conclusion, simply including the measurement matrix into end-to-end schemes like \eqref{unrolling} does not respect important constraints and therefore limits its practicality.

\subsection{Gumbel Reparametrizations}

Backpropagating gradients through immensely large computational graphs has shown to be highly effective in the training of deep neural networks with hundreds of layers \cite{resnet152} and billions of learnable parameters \cite{gpt3}. However, when discrete nodes are included in the computational graphs, gradients cannot be computed directly and have to be estimated. The Gumbel reparametrization is a method for gradient-based learning of a probability distribution underlying discrete sampling processes.\footnote{From a signal processing point of view, this can be related to derandomization, i.e., finding a probability measure on sensing matrices with low entropy, e.g., see \cite{Jung:frontiers18}.}
Formally, we consider a computational graph including a discrete random variable~$v$ taking values in $\{1, \dots, a\}$ (in one-hot encoding, i.e., every element in $\{1, \dots, a\}$ is represented as a different unit vector in $\R^a$) for some $a \in \mathbb{N}$, where its unnormalized log-probabilities are denoted by $\varphi = [\varphi_1, \dots, \varphi_a]^T \in \mathbb{R}^a$. The value of $v$ is then passed through a deterministic, differentiable function~$f$. 
The goal is to compute gradients with respect to the probability parameters $\varphi$.
As~$v$ is discrete, it is not possible to directly backpropagate through~$v$. However, the gradient with respect to $\varphi$ can be estimated via sampling from~$v$:

\begin{equation*}
	\nabla_\varphi \E_v \Big[f(v)\Big].
\end{equation*}

A simple yet effective way to sample from such a categorical random variable is to use the Gumbel-Max trick \cite{gumbelmax, gumbelmax2}, according to which a realization of $v$ can be obtained by

\begin{equation}
	\label{gumbelmaxeq}v_i = \text{one-hot}\big(\argmax_{j \in \{1, \dots, a\}} (g_j + \varphi_j)\big)_i, \quad i = 1,\dots, a,
\end{equation}

where $g_j \sim \text{Gumbel}(0, 1)$ follows a Gumbel distribution, which is equivalent to a twice negative-log transformed uniform distribution, i.e., $g_j = -\log ( - \log(u_j))$ with $u_j \sim\nobreak U(0, 1)$.

The Gumbel-softmax trick \cite{gumbel}, also known as the Concrete distribution \cite{concrete}, replaces the argmax operator in $\eqref{gumbelmaxeq}$ by a softmax operator (with temperature $\tau$), which ensures differentiability everywhere:

\begin{equation}\label{gumbelsoftmaxeq}
	v_i = \frac{\exp((g_i + \varphi_i) / \tau)}{\sum_{j=1}^{l}\exp((g_j+ \varphi_j) / \tau)}, \quad i = 1,\dots, a.
\end{equation}

Note that in the limit $\tau \rightarrow 0$, the Gumbel-softmax trick based on \eqref{gumbelsoftmaxeq} reduces to the non-differentiable argmax in~\eqref{gumbelmaxeq}.
This modification allows for the backpropagation through discrete random variables by adding element-wise i.i.d.~noise, outperforming previous approaches to estimating the gradient of discrete nodes, such as the straight-through estimator \cite{straightthrough}, or score-function-based estimators \cite{reinforce, nvil} in a series of supervised learning tasks. For $\tau > 0$, the Gumbel-softmax is a continuous relaxation of the (one-hot encoded) discrete random variable. In cases where true discreteness is needed, it is possible to use the argmax operator in the forward pass and the softmax in the backward pass of backpropagation. This strategy is known as the straight-through Gumbel softmax estimator. It can be directly extended to taking multiple samples without replacement from a categorical distribution, namely by selecting the top-$K$ values instead of only the largest one. This was first highlighted in a blog post by Vieira~\cite{blogpost}, connecting Gumbel reparametrizations to weighted reservoir sampling.

Gumbel reparametrizations have found various applications in the field of machine learning, ranging from learned discrete representations \cite{gumbel, learned_discrete1}, over graph generation \cite{gggan}, to neural architecture search \cite{nas}. Recently, they have been also applied in the context of compressive magnetic resonance imaging \cite{huijben2019deep,huijben2020learning}, where a single Gumbel top-$K$ operation is employed to select a fixed number of rows from a Fourier matrix; see also \cite{bahadir2019adaptive,bahadir2020deep}.
This can be seen as a special case of the GLODISMO framework (see Section~\ref{section:method}), when learning only a single global mask.
The aforementioned works rely on off-the-shelf feedforward neural networks for recovery, which is a notable difference to our unrolling-based approach.\footnote{Note that this can be viewed as a recurrent neural network in \eqref{unrolling}, since the matrix $\Phi$ appears in every iteration (layer) of the unrolled algorithm, cf.~\eqref{proximal}.}
In fact, the use of unrolling comes along with several important practical advantages.

First, unrolled methods like the variants of LISTA can be designed in a very memory-efficient fashion, due to explicit control over the learnable components of the unrolled iterative (convex) scheme, e.g., see \cite{alista,na-alista}.
This allows overcoming the limitations of memory-intensive feedforward neural networks when subjected to additional constraints, e.g., the deployment on mobile hardware.

Second, it has been demonstrated that unrolled methods are capable of much more accurate and robust reconstructions, as they take the forward model of the inverse problem explicitly into account, e.g., see \cite{genzel-robustness,genzel-accuracy,leuschner21,hsqdsr19,sp21}.
In a similar vein, the training of unrolled networks is typically also less prone to overfitting than post-processing feedforward neural networks.

\section{Method}
\label{section:method}

In this section, we present our method GLODISMO (Gradient-based Learning of \mbox{DIscrete} Structured Measurement Operators).
The main objective is to combine the unrolling approach of \eqref{unrolling} with learning a measurement operator $\Phi$:

\begin{equation}\label{glodismo}
	\min_{\theta, \Phi} \ \E_{x, z} \Big[\L(f_\theta(\Phi x + z), x)\Big].
\end{equation}

Our key idea is to impose discrete structural constraints on~$\Phi$ by Gumbel reparametrizations to enable gradient estimation.

More specifically, we learn a probability distribution parametrized by a learnable parameter $\varphi \in \R^{m \times n}$, from which the (discrete and structured) matrix-valued measurement operator $\Phi \in \R^{m \times n}$ can be drawn.
To this end, we consider the index set $\I := \{1, \dots, m\} \times \{1, \dots, n\}$ and let $\P(\I) = \{I_1, \dots, I_l\}$ be a partition\footnote{That means the sets $I_1, \dots, I_l$ are non-empty and pairwise disjoint such that $\bigcup_{i \in \{1, \dots, l\}} I_i=\I$.} of~$\I$, whose purpose is to capture the structural constraints on $\Phi$.
As such, $\Phi$ follows a joint distribution over random variables supported on the index subsets $I_1, \dots, I_l$, each of which is independently obtained from a Gumbel reparametrization.
At this point the parameter $\varphi$ comes into play: The notation $\varphi[I_i]$ refers to those entries of $\varphi$ indexed by $I_i$. 
To obtain~$\Phi$, for each $I_i$, we add element-wise i.i.d.~Gumbel noise to $\varphi[I_i]$ and then select the indices corresponding to the \mbox{top-$d_i$} values.
This procedure is used in the forward pass to sample~$\Phi$ and to compute the loss in \eqref{glodismo}.
In the backward pass, we use the gradient of the softmax with temperature $\tau$ instead of the hard \mbox{top-$d_i$}. 
We refer to the above procedure as GLODISMO.
A pseudo-code implementation is provided in Algorithm~\ref{alg:algorithm}, assuming a software framework capable of automatic differentiation.
For the sake of simplicity, the described method is limited to learning binary matrices, but an extension to larger alphabets is straightforward.

\begin{algorithm}[t]
	\caption{\textbf{(GLODISMO)} Learning a binary matrix with $d_i$ ones per set~$I_i$ of the partition.}
	\label{alg:algorithm}
	\textbf{Input:} signal $x$ (training data), temperature $\tau$, top-$K$-keeps $d_1, \dots, d_l \in \mathbb{N}$, differentiable reconstruction algorithm $f_\theta :\nobreak \R^m \to \R^n$, index partition $\P(\I)= \{I_1, \dots,  I_l\}$, perturbing noise distribution $\mathcal{Z}$ \\
	\textbf{Learnable Parameters:} Parameters of measurement matrix $\varphi \in \R^{m \times n}$, parameters of reconstruction algorithm~$\theta$
	\begin{algorithmic}[1] 
		\STATE $G \sim_{i.i.d.} \text{Gumbel}(0, 1)^{m \times n}$
		\FOR {$i \in \{1, \dots, l \}$}
		\STATE logits := $(\varphi[I_i] + G[I_i]) / \tau$\\
		\STATE probs := softmax(logits)\\
		\STATE hard := topk(probs, $d_i$)\\
		\STATE $\Phi[I_i]$ := hard.detach() + probs - probs.detach()\\
		\ENDFOR
		\STATE $z \sim_{i.i.d.} \mathcal{Z}^{m}$\\
		\STATE  $y := \Phi x + z$\\
		\STATE  loss $ := \L(f_\theta(y), x)$\\
		\STATE  loss.backward()
	\end{algorithmic}
\end{algorithm}

As indicated above, the index partition $\P(\I)$ is used to impose structural constraints on the measurement matrix~$\Phi$.
While a proper choice of $\P(\I)$ is problem-specific, it is often naturally given by the rows or columns of $\Phi$, and the number of selected elements $d_i$ is equal for each $i \in \{1,\dots,l\}$ (see Section~\ref{experiments:section} for examples).
In such cases, vectorized, and therefore computationally efficient softmax and top-$K$ operations over one of the axes can be applied.

In Algorithm~\ref{alg:algorithm}, $\varphi \in \R^{m \times n}$ has $m \cdot n$ learnable parameters, which must be stored in memory during training time. This is identical to the cost of learning a dense sensing matrix without constraints. At test time, a single mask is sampled after adding Gumbel noise and performing the top-$K$, and then kept fixed. This means that there is no additional computational cost compared to using a conventional (randomly chosen, but fixed) matrix satisfying the constraints. Hence, our method is feasible for training and comes at no additional expenses during testing or deployment.

Note that the binary matrix computed in Algorithm~\ref{alg:algorithm} can be used like any other parameter in a framework with automatic differentiation, since the gradient with respect to~$\varphi$ is well-defined.
In particular, this allows for simultaneous optimization of~$\Phi$ and other tunable parameters~$\theta$, which is an important advantage over discrete optimization techniques.
More generally, GLODISMO should be seen as an extendable learning framework rather than a rigid algorithm.
For example, Algorithm~\ref{alg:algorithm} can be easily modified to learn any measurement operator that is constructed by differentiable transforms of one or even multiple binary matrices (which are structured in the sense that their entries are subselected from specific partitions).
We demonstrate this flexibility in the context of fast transforms (see Appendix~\ref{appendix:fasttransforms}) and learning super-pixel masks for single pixel imaging (see Appendix~\ref{appendix:superpixel}).
A full evaluation of potential generalizations is beyond the scope of this paper and left to future research.

\section{Experiments}
\label{experiments:section}

In this section, we demonstrate the effectiveness of GLODISMO in a selection of prototypical signal recovery applications. We refer the reader to Appendix~\ref{appendix:implementation_details} for more details on the implementation of all considered methods.

\paragraph*{Baselines} We compare our method to random matrices (fixed after a single random draw), which is a standard baseline for all considered problem setups.
Moreover, we benchmark against two baseline discrete optimization algorithms: a greedy approach as well as Simulated Annealing \cite{simulatedannealing, simulatedannealing2}, referred to as ``Greedy'' and ``SimAn'' in our experiments, respectively.
Note that both methods are incompatible with reconstruction algorithms that include any (real-valued) learnable parameters.

\paragraph*{Error metrics} As common in many signal recovery tasks, we report the normalized mean squared error (NMSE) and normalized mean absolute error (NMAE) between the ground truth signal $x$ and the estimated reconstruction $\hat{x}$, which are defined as (in dB-scale):

\begin{align*}
\text{NMSE} (x, \hat{x})&:= 10 \log_{10}\Big(\frac{\mathbb{E}_{x}[\|\hat{x}- x\|_2^2]}{\mathbb{E}_{x}[\|x\|_2^2]}\Big), \\
\text{NMAE}(x, \hat{x}) &:= 10 \log_{10}\Big(\frac{\mathbb{E}_{x}[\|\hat{x}- x\|_1]}{\mathbb{E}_{x}[\|x\|_1]}\Big).
\end{align*}

\subsection{Application: Single Pixel Imaging}
\label{singlepixel:section}

\begin{figure}[b]
	\begin{center}
		\centerline{\includegraphics[width=0.8\columnwidth]{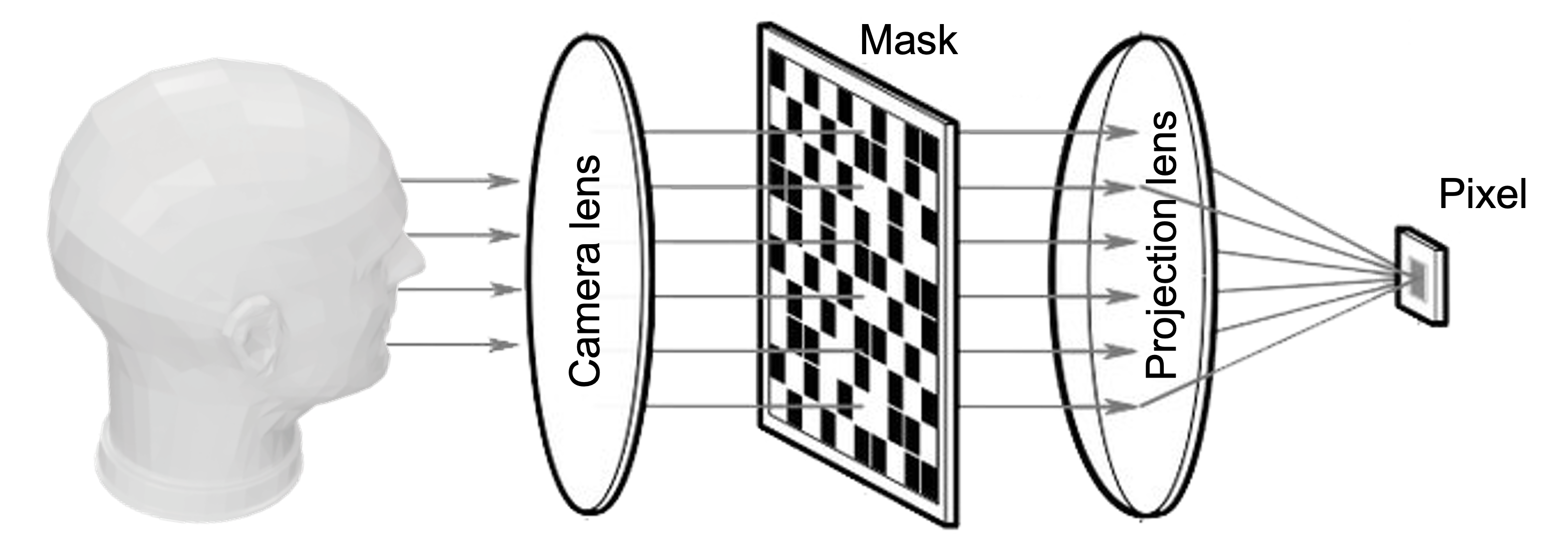}}
		\caption{Schematic visualization of a single-pixel imaging setup; image adapted from \cite{bacca2019}.}
		\label{singlepixel:example}
	\end{center}
\end{figure}

The fact that conventional visible light cameras with millions of pixels are cheaply available stems from the fortunate coincidence that the wavelengths of visible light are in the same order of magnitude as the wavelengths that silicon responds to \cite{singlepixelreport}. For imaging beyond visible light, cameras composed of many pixels may be prohibitively expensive due to the difficulty in manufacturing such pixels. A remarkable application of compressed sensing is single pixel imaging \cite{singlepixel}, in which the original image is reconstructed from compressive measurements acquired by a spatial light modulator or a digital micromirror device that collects light onto a single pixel using a series of masks. For each of the $m$ measurements, the pixel intensity is measured while using a distinct mask. Mathematically, this process can be expressed as observing the unknown (vectorized) image $x \in \R^n$ through a binary measurement matrix~$\Phi \in \R^{m \times n}$, and then using a reconstruction algorithm to estimate the original image from $m$ sums of pixels. As natural images are often sparse with respect to a certain transform domain, compressed sensing can be employed for the reconstruction. An example of a single-pixel imaging setup is visualized in Figure~\ref{singlepixel:example}.
While theoretical results highlight that random masks have favorable reconstruction properties for generic sparse signals, this may not be the case when the physics and hardware constraints of the measurement process or additional image structures are taken into account. Examples are the modeling of diffraction effects or implicit signal structure that is extracted in a data-driven fashion.

GLODISMO is well suited to incorporate such aspects. 
For our case study, we partition the indices of $\varphi \in \R^{m \times n}$ into the set of row vectors and use Algorithm~\ref{alg:algorithm} to learn adaptive masks for the MNIST dataset ($n = 784$) \cite{lbbh98}. The squared error $\L(\hat{x}, x) =\|\hat{x} - x\|_2^2 $ is used as loss function during training, and bi-orthogonal~2.2 wavelets with one level as sparsifying transform. We unroll IHT, which has no additional learnable parameters,\footnote{Following \cite{iht}, the step size is fixed as 1. Here, hard thresholding in IHT keeps $50$ components in the wavelet basis.} and \mbox{NA-ALISTA},\footnote{Our implementation of NA-ALISTA employs support selection during thresholding as in \cite{listacpss} but simply replaces the analytic computation of the weights proposed in \cite{alista} by its adjoint. Otherwise, the analytic computation would have to be re-computed any time a new $\Phi$ is used, ergo at every step.} in which an LSTM-network \cite{lstm} is employed to adaptively predict step-sizes and thresholds. 
In a first experiment, we demonstrate that GLODISMO enables much better image reconstructions from far fewer measurements. For this, we fix the number of ones per row (i.e., the number of "on" pixels per mask) to 32. We run IHT and \mbox{NA-ALISTA} for $T=20$ iterations and compare the reconstruction performance for random pixel masks and those learned via GLODISMO. The number of measurements is varied from 10 to 500. Additive i.i.d. Gaussian noise with a signal-to-noise ratio of 40dB is added to the measurement vector~$y$.

\begin{figure}
	\begin{center}
		\centerline{\includegraphics[width=0.8\columnwidth]{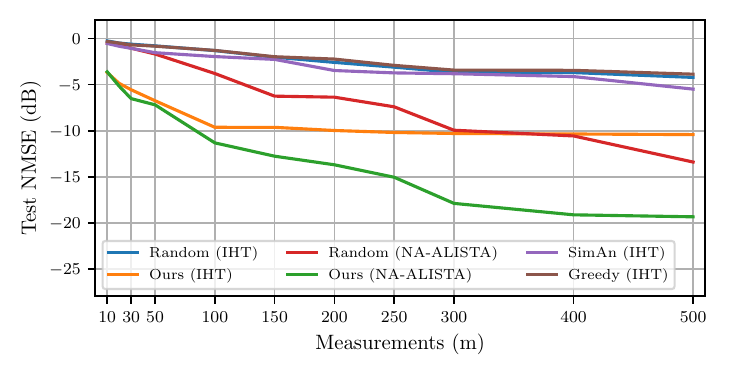}}
	\caption{NMSE of reconstruction with single pixel imaging masks as a function of the number of measurements $m$.
	}
	\label{singlepixel:pixelconvergence_varym}
	\end{center}
\end{figure}

The results are reported in Figure~\ref{singlepixel:pixelconvergence_varym}. With only 10 measurements, IHT with a learned $\Phi$ is able to match the performance of IHT with a random $\Phi$ of 200 measurements. As IHT itself has no learnable parameters, this difference must stem from the learned mask.
Both the greedy and Simulated Annealing baselines lead to only marginally better results than random matrices. As the hard-thresholding in IHT removes all but the top 50 absolute values of the reconstruction (in the wavelet basis), it saturates at around -10dB NMSE for learned $\Phi$, which is essentially perfect reconstruction of the top 50 wavelet coefficients under 40dB Gaussian noise. On the other hand, \mbox{NA-ALISTA}, which is based on soft-thresholding, is able to reach a significantly higher reconstruction quality. Similarly to IHT, \mbox{NA-ALISTA} with a learned mask significantly outperforms its counterpart with a random mask. Overall, Figure~\ref{singlepixel:pixelconvergence_varym} highlights that both learning~$\Phi$ and learning the parameters of the reconstruction algorithm enable more accurate and faster reconstructions; accordingly, this enables a significant reduction of the number of required measurements.

\begin{figure*}[t]
	\begin{center}
		\centerline{\includegraphics[width=0.8\columnwidth]{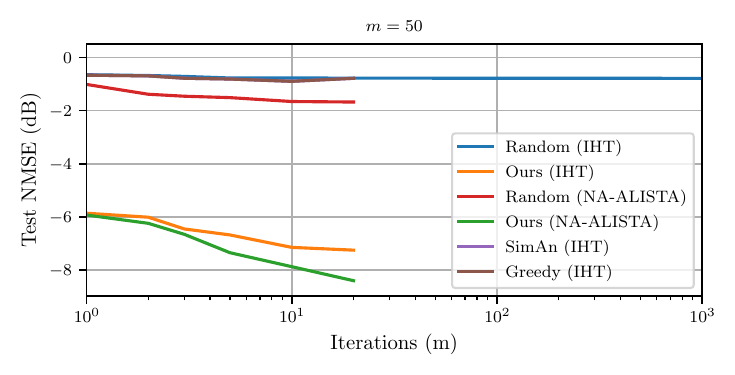} \qquad\qquad
		\includegraphics[width=0.8\columnwidth]{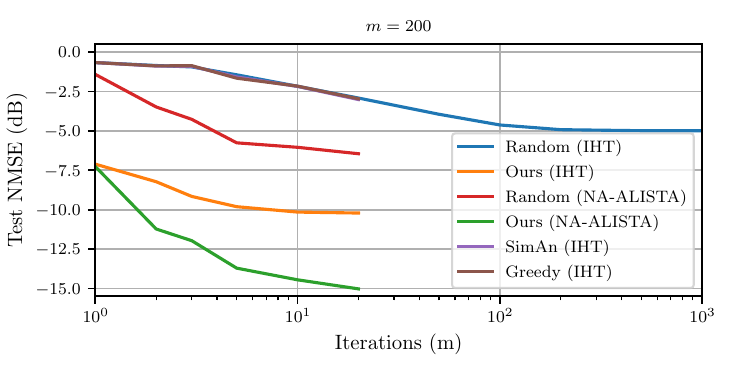}}
		\caption{NMSE of image reconstruction with single pixel imaging masks as a function of the number of iterations $t$ for $m=50$ (left) and $m=200$ measurements (right).}
		\label{singlepixel:pixelconvergence_varyk}
	\end{center}
\end{figure*}

The effectiveness of GLODISMO is further underpinned when fixing the number of measurements and instead evaluating the improvement in convergence speed. For $m=50$ and $m=200$ measurements, we compare the convergence of the aforementioned algorithms in Figure \ref{singlepixel:pixelconvergence_varyk}. Even after a single iteration of using a learned $\Phi$ with either IHT or \mbox{NA-ALISTA}, the reconstruction is better than a random~$\Phi$ with a thousand iterations of IHT or 20 iterations of \mbox{NA-ALISTA}. Furthermore, a learned $\Phi$ leads to a much higher accuracy after only a few iterations when compared to IHT with a random $\Phi$ after thousands of iterations. This also holds true for $m=200$, but \mbox{NA-ALISTA} outperforms IHT for learned $\Phi$. Again, both the greedy and Simulated Annealing baseline are only insignificantly better than random matrices. These results verify that GLODISMO can be used seamlessly within the algorithm unrolling framework to speed up convergence of signal recovery.

\begin{figure}[t]
	\begin{center}
		\begin{subfigure}{0.95\columnwidth}
			\includegraphics[width=0.24\columnwidth]{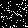}
			\includegraphics[width=0.24\columnwidth]{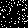}
			\includegraphics[width=0.24\columnwidth]{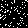}
			\includegraphics[width=0.24\columnwidth]{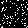}
			\caption{Random Masks\vspace{.25\baselineskip}}
		\end{subfigure}
		\begin{subfigure}{.95\columnwidth}
			\includegraphics[width=0.24\columnwidth]{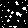}
			\includegraphics[width=0.24\columnwidth]{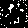}
			\includegraphics[width=0.24\columnwidth]{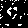}
			\includegraphics[width=0.24\columnwidth]{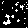}
			\caption{Learned Masks\vspace{.25\baselineskip}}
		\end{subfigure}
		\begin{subfigure}{0.95\columnwidth}
			\includegraphics[width=\columnwidth]{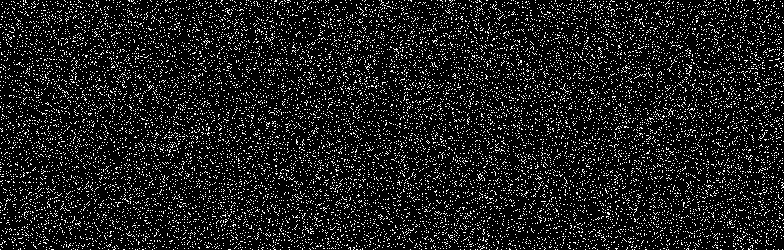}
			\caption{Random $\Phi$\vspace{.25\baselineskip}}
		\end{subfigure}
		\begin{subfigure}{0.95\columnwidth}
			\includegraphics[width=\columnwidth]{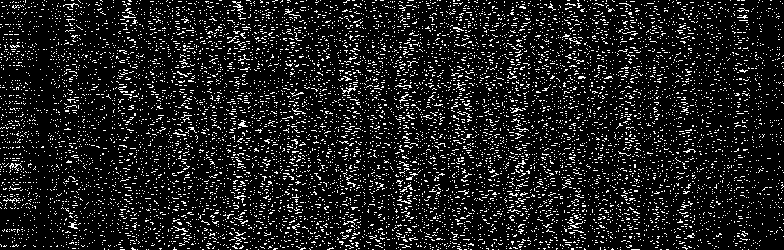}
			\caption{Learned $\Phi$\vspace{.25\baselineskip}}
		\end{subfigure}
		\begin{subfigure}{.95\columnwidth}
			\includegraphics[width=0.24\columnwidth]{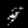}
			\includegraphics[width=0.24\columnwidth]{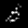}
			\includegraphics[width=0.24\columnwidth]{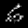}
			\includegraphics[width=0.24\columnwidth]{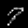}
			\caption{Reconstructed (random $\Phi$)\vspace{.25\baselineskip}}
		\end{subfigure}
		\begin{subfigure}{.95\columnwidth}
			\includegraphics[width=0.24\columnwidth]{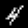}
			\includegraphics[width=0.24\columnwidth]{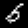}
			\includegraphics[width=0.24\columnwidth]{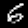}
			\includegraphics[width=0.24\columnwidth]{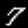}
			\caption{Reconstructed (learned $\Phi$)\vspace{.25\baselineskip}}
		\end{subfigure}
		\caption{Single pixel imaging masks for $m=250$, where $d=90$ ones per row are selected (i.e., ones per mask). The reconstructions are obtained by $T=20$ IHT-iterations.}
		\label{singlepixel:pixel_masks}
	\end{center}
\end{figure}

Finally, to visualize how changing the mask influences the reconstruction quality beyond a quantitative metric like NMSE, Figure \ref{singlepixel:pixel_masks} shows a random~$\Phi$ and a learned~$\Phi$ (in combination with unrolled IHT), as well as individual reconstructions from the test set. The four images in subfigure~(a) correspond to the top four rows of the matrix which is displayed in subfigure~(c) (rows resized to image shape), i.e., the first four light patterns that are projected onto the scene. No structure can be seen in the matrix or the masks, and the samples reconstructed with 20 IHT-iterations in subfigure~(e) are of poor quality. Figure~\ref{singlepixel:pixel_masks} also displays the learned counterparts in subfigures~(b), (d), and (f), respectively. The four shown masks clearly exhibit additional structure, and so does the full matrix $\Phi$. The digits of the reconstructed samples are clearly readable.

\paragraph*{Further experiments} We have conducted further experiments on the single-pixel imaging setup to highlight possible extensions of GLODISMO within the automatic differentiation framework. More specifically, we show that GLODISMO can be used to learn fast-transform matrices in Appendix~\ref{appendix:fasttransforms} as well as super-pixel masks in Appendix~\ref{appendix:superpixel}.

\subsection{Application: Compressed Sensing with Left-$d$-Regular Graphs}

\begin{figure}[ht!]
	\begin{center}
		\centerline{\includegraphics[width=0.5\columnwidth]{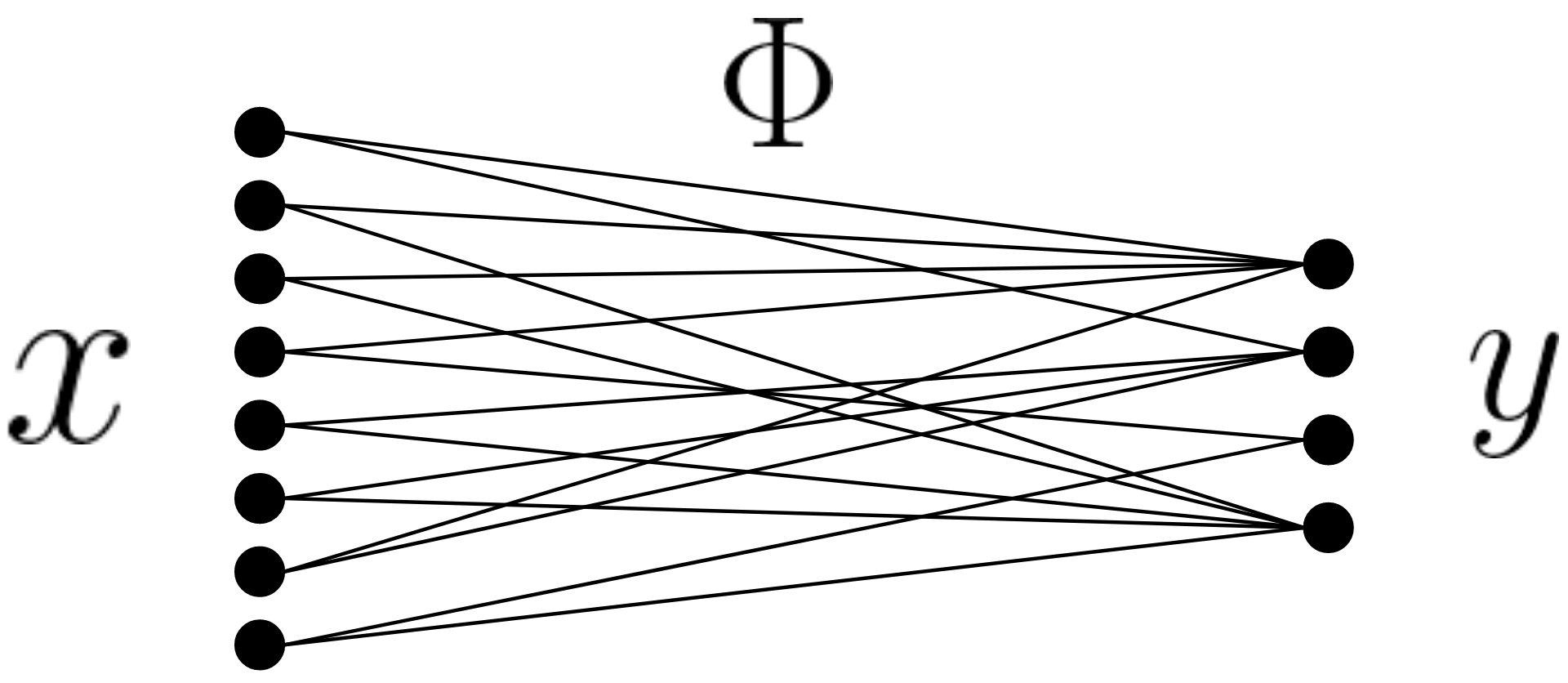}}
		\caption{Example of a linear measurement process using a left-$d$-regular bipartite graph with $d = 2$, where the signal space is 8-dimensional and the measurement space 4-dimensional; image adapted from \cite[p.~436]{foucart2013invitation}.}
		\label{expander:example}
	\end{center}
\end{figure}

\begin{figure}[ht!]
	\begin{center}
		\centerline{\includegraphics[width=.8\columnwidth]{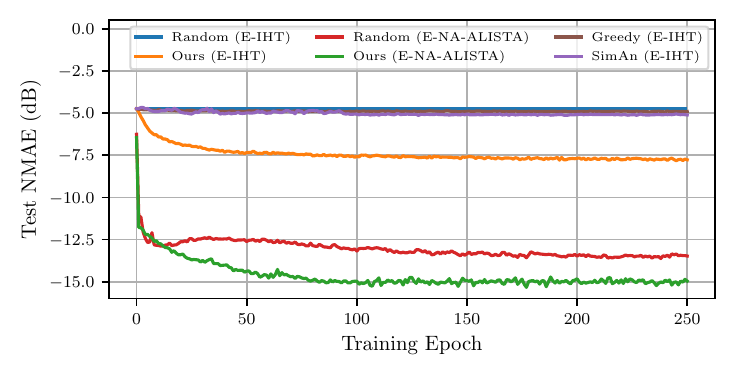}}
		\caption{NMAE of compressed sensing with left-$d$-regular graphs on synthetic data as a function of the training progress.}
		\label{expander:loss_epoch}
	\end{center}
\end{figure}

\begin{figure}[ht!]
	\begin{center}
		\centerline{\includegraphics[width=.8\columnwidth]{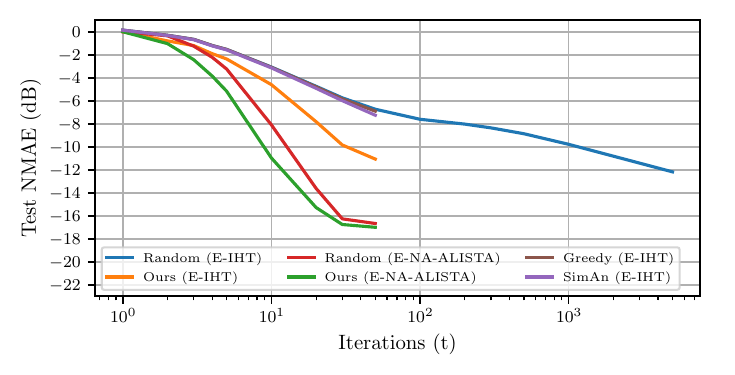}}
		\caption{NMAE of compressed sensing with left-$d$-regular graphs on synthetic data as a function of the number of iterations $t$.}
		\label{expander:convergence}
	\end{center}
\end{figure}

Another important class of measurement matrices suitable for compressed sensing is given by adjacency matrices of lossless expanders \cite{foucart2013invitation}. In this setting, $\Phi$ corresponds to a 0-1 adjacency matrix of a bipartite graph, connecting the signal~$x$ (left vertices) to the measurements~$y$ (right vertices). A graph is called left-$d$-regular if every left vertex has exactly $d$ connected right vertices. In matrix notation, this translates into $\Phi$ having exactly $d$ ones per column. An example is visualized in Figure \ref{expander:example}. Extremely large random left-$d$-regular graphs with  $d \in O(n/s)$ and $m \in O(s \log(n/s))$ satisfy the restricted expansion property with high probability, which is a sufficient recovery criterion \cite{foucart2013invitation}.
However, for smaller values of~$m$ and~$n$, such random graphs are unlikely to enjoy this property and therefore may not be suited for compressed sensing.
In this regime, we can employ GLODISMO to learn left-$d$-regular graphs. Invoking Algorithm~\ref{alg:algorithm}, we partition the entries of the measurement matrix into its column vectors, and select $d=7$ ones per column. As unrolled method, we consider Iterative Hard Thresholding for expanders (\mbox{E-IHT}), see \cite{foucart2013invitation}:
\begin{equation*}
	\xh^{(t+1)} = \mathcal{H}_s \Big(\xh^{(t)} + \mathcal{M}\big(y - \Phi \xh^{(t)} \big) \Big).
\end{equation*}
Here, $\mathcal{H}_s: \R^{n} \rightarrow \Sigma_s^n$ denotes the hard thresholding operator, which sets all but the $s$ largest absolute coefficients to~$0$. The nonlinear function $\mathcal{M}: \R^m \to \R^n$ denotes the median operator associated with $\Phi$, i.e., the $j$-th coefficient of $\mathcal{M}(y)$ is given by $\text{median}\{ y_i \, : \, i \in R(j) \}$, where $R(j)$ is the set of right vertices connected to the left vertex $j$ (which are $d$-many for left-$d$-regular graphs).
Similarly to \mbox{E-IHT}, we adapt \mbox{NA-ALISTA} to expanders by replacing the adjoint~$\Phi^T$ in each iteration with the median operator~$\mathcal{M}$, which we call \mbox{E-NA-ALISTA}. As common in expander theory, we use $\mathcal{L}(x, \hat{x}) = \|x - \hat{x}\|_1$ as loss function. The experiments in this section are conducted with heavy-tailed noise (student t-distributed with 1 degree of freedom) and a signal-to-noise ratio of 40dB.

We employ \mbox{E-IHT} and \mbox{E-NA-ALISTA} with $m=250$ measurements on synthetic sparse vectors with $n=784$ and an expected sparsity of $s=40$. The support of the synthetic data is generated via i.i.d.~Bernoulli random variables, while the non-zero coefficients are normally distributed. Figure~\ref{expander:loss_epoch} reports the NMAE on the test set as the training progresses; note that \mbox{E-IHT} itself has no learnable parameters. GLODISMO clearly improves the reconstruction performance for both \mbox{E-IHT} and \mbox{E-NA-ALISTA}. This is remarkable from a compressed sensing perspective, since there is no signal structure beyond generic sparsity, and yet the learned matrices empirically outperform their random counterparts.

To evaluate if GLODISMO can speed up solving inverse problems with left-$d$-regular graphs, we report the NMAE as a function of the algorithm iterations in Figure~\ref{expander:convergence}. \mbox{E-IHT} with a random left-$d$-regular graph converges much more slowly than in the case of learning: the former takes about 2000 iterations to reach the same NMAE level as the latter with only 50 iterations. \mbox{E-NA-ALISTA} generally performs better than \mbox{E-IHT} and can be also further accelerated by GLODISMO.
Similar to single pixel imaging, we observe in both Figures~\ref{expander:loss_epoch} and~\ref{expander:convergence} that the greedy and Simulated Annealing baselines only slightly improve upon random designs.

\begin{figure}
	\begin{center}
		\centerline{\includegraphics[width=.8\columnwidth]{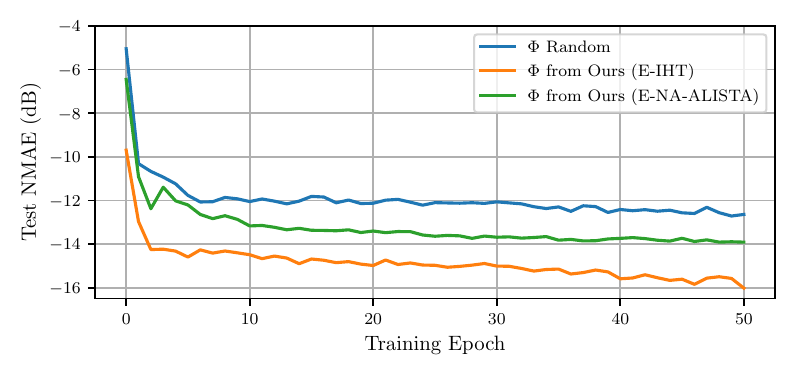}}
		\caption{Comparison of training \mbox{E-NA-ALISTA} with a fixed left-$d$-regular graph, which is randomly drawn (blue), previously learned by \mbox{E-IHT} with $T=20$ iterations (orange), and previously learned by \mbox{E-NA-ALISTA} with $T=20$ iterations (green).}
		\label{expander:generalizationa}
	\end{center}
\end{figure}

\begin{figure}
		\begin{center}
			\centerline{\includegraphics[width=.8\columnwidth]{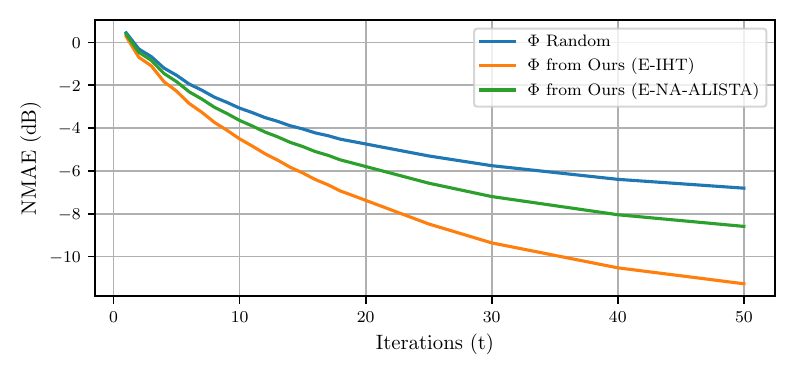}}
			\caption{Comparison of \mbox{E-IHT} (with varying number of iterations~$t$) using a fixed left-$d$-regular graph, which is randomly drawn (blue), previously learned by \mbox{E-IHT} with $T=20$ iterations (orange), and previously learned by \mbox{E-NA-ALISTA} with $T=20$ iterations (green).}
			\label{expander:generalizationb}
		\end{center}
\end{figure}

Finally, an interesting question is whether the measurement matrices learned by GLODISMO overfit to the unrolled algorithm in use. To this end, we evaluate whether a matrix learned by \mbox{E-IHT} also works well for \mbox{E-NA-ALISTA} and vice versa (in the same setup as before). The results are reported in Figures~\ref{expander:generalizationa} and~\ref{expander:generalizationb}. We find that using a matrix previously learned by \mbox{E-IHT} as a fixed matrix for \mbox{E-NA-ALISTA} leads to a slightly better performance than learning it directly with \mbox{E-NA-ALISTA}. In the other direction, using a matrix learned by \mbox{E-NA-ALISTA} for \mbox{E-IHT} works better than a random matrix, but it is outperformed by a matrix specifically learned by \mbox{E-IHT}. 
These observations appear plausible to us, since \mbox{E-NA-ALISTA} learns the measurement operator and algorithm parameters simultaneously, whereas \mbox{E-IHT} itself has no tuning parameters.

\subsection{Application: Pooling Matrices for Group Testing}

Another prototypical inverse problem with discrete structured measurement operators is posed by group testing, which is a technique that can be used to reduce the number of tests to identify infectious diseases. Conceptually, the measurement matrix describes which specimen is pooled into which test, while the actual recovery problem is conveniently addressable by Non-Negative Least Absolute Deviation (NNLAD) algorithms \cite{petercovid19}.
An analysis of GLODISMO in this setup is presented in Appendix~\ref{appendix:grouptesting}.

\section{Conclusion \& Future Work}

In this work, we have proposed GLODISMO, an efficient and extendable method for learning discrete structured measurement operators for signal recovery, based on a fusion of unrolled optimization and Gumbel reparametrizations. 
The effectiveness and flexibility of our approach has been empirically demonstrated in several prototypical applications, thereby significantly outperforming discrete optimization baselines and random matrix designs. There are many more potential extensions and applications of GLODISMO, which could be explored in future work, some of which are mentioned in the following:
\begin{itemize}
\item 
	\emph{Pre-trained models.} For certain inverse problems, there exist powerful pre-trained reconstruction networks, e.g., in magnetic resonance imaging (MRI) \cite{kno+20b,muc+20}.
	As long as a given model is implemented in an end-to-end fashion (i.e., allows for backpropagation), it could be used as pre-initialized model for GLODISMO. 
	In the case of MRI, this can be particularly useful if one would like to learn new sampling masks for different hardware, while the data distribution remains (approximately) the same.
\item 
	\emph{Non-linear inverse problems.} Our methodology could be extended to inverse problems with non-linear forward operators like phase-retrieval.
	A well-known obstacle is that the corresponding variational formulations typically become non-convex, and iterative schemes require a good initialization to converge.
	This might especially harm the effectiveness of unrolled methods.
	One possible remedy would be to compose the unrolled scheme with a neural network that was pre-trained to provide good initializations (for some initial measurement matrix).
	This pipeline is then learned end-to-end with GLODISMO.
\item 
	\emph{Different low-dimensional structures.} It is certainly possible to apply GLODISMO to signal structures beyond plain sparsity, e.g., sparse vectors with additional fine-structure for which specialized denoising functions are required instead of plain thresholding~\cite{Musa:icassp21}.
	Similarly, measurement operator learning is also applicable to matrix recovery problems like unrolled ``sparse plus low-rank'' decompositions \cite{SanchezPastor:sensors:2021}.
\item
	\emph{Large-scale simulations and classification.} It is evident to evaluate our methodology in a large-scale setting, with much higher ambient dimensions (both in the signal and measurement domain) and larger datasets. On the other hand, future research could also involve an extension to compressive classification \cite{compressive1, compressive2}, where the measurement operator would be optimized to classify signals based on the measurements, instead of reconstructing them.
\end{itemize}


\appendices

\section{Implementation Details}
\label{appendix:implementation_details}

In our experimental setup,\footnote{The full code of our implementation is available at \href{https://github.com/josauder/glodismo}{\texttt{https://github.com/josauder/glodismo}}} we optimize $\Phi$ and $\theta$ using the Adam optimizer \cite{adam} with $\beta_1=\nobreak 0.9$, $\beta_2=0.999$ (default in PyTorch \cite{pytorch}) and a fixed learning rate of 0.0002. We use mini-batches of size~512 training samples. One epoch of training with synthetic data is comprised of 50 000 samples, while the test set with 10 000 samples is kept fixed and not seen during training. The parameters $\varphi$ are initialized using a standard Gumbel distribution. We find that when learning both $\Phi$ and $\theta$, training is more stable when rescaling the Gumbel noise by a factor of 0.001, both for initialization as well as for the Gumbel reparametrizations during training. We keep the softmax temperature in all Gumbel reparametrizations fixed at $\tau=1$, which has been demonstrated to work well in practice \cite{gumbel}. For all evaluated algorithms, the learned $\Phi$ is always initialized with the same random seed, which also yields the random but fixed $\Phi$ after the top-$K$ operation. Before each training run, the optimal scalar scaling factor of the measurement operator is determined via a grid search using the initial $\Phi$. When training on MNIST, the previously determined optimal scalar is multiplied by 0.9 to prevent gradient explosion as more pixels move to the center of the masks.

\paragraph*{Discrete optimization baselines} In each optimization step, the baseline discrete optimization algorithms propose neighbors to the current $\Phi$, which are either accepted or rejected. We benchmark against a greedy baseline and a Simulated Annealing (SimAn) \cite{simulatedannealing, simulatedannealing2} baseline. Let the shorthand $\L_\Phi$ denote the loss over a randomly sampled mini-batch using $\Phi$ as a measurement operator. The greedy algorithm only accepts neighbors $\Phi^\prime$ that strictly decrease the loss, i.e., when $\L_{\Phi^\prime} < \L_\Phi$. The SimAn baseline can also accept neighbors that increase the loss, with the rule becoming stricter as training progresses. More precisely, SimAn accepts a neighbor $\Phi^\prime$ if it strictly decreases the loss or if $\exp\big((\L_\Phi - \L_{\Phi^\prime})/\tau\big) < u$, where $u \sim U(0,1)$ is drawn from a uniform distribution and $\tau$ is a temperature parameters. The acceptance probability depends crucially on $\tau$: initially $\tau$ is chosen larger, such that SimAn explores solutions far from its initial $\Phi$. As training progresses, $\tau$ is steadily decreased. We manually tuned the initial $\tau$ to accept about 80\% of all neighbors in the first epoch and the decay rate such that changes are accepted for at least 100 training epochs. For single pixel imaging this revealed an initial $\tau$ of 0.0012 and a temperature decay factor of 0.9997. For left-$d$-regular graphs, a slightly larger $\tau$ of 0.003 and a temperature decay of 0.9998 was determined to work well. We found that increasing the batch size 10-fold (to stabilize the quantities $\L_{\Phi}$ and $\L_{\Phi^\prime}$) did not have a significant effect on the performance of the baseline algorithms.

Neighbors to the current $\Phi$ are chosen as follows: a random subset of the partition defining the structure is chosen. In this subset, the position of a random one and a random zero are swapped. For example, in the case of single-pixel imaging, this amounts to selecting a random row, and swapping a random one with a random zero within that row. This procedure for drawing neighbors ensures that the neighbors always satisfies the constraints. It also fulfills a notion of being the `smallest' step away from $\Phi$.

\section{Supplemental Experiments}

\begin{figure}[H]
\begin{center}
\includegraphics[width=0.45\columnwidth]{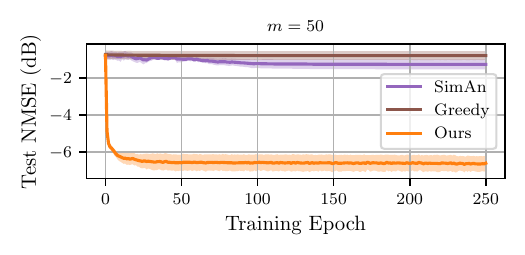}
\includegraphics[width=0.45\columnwidth]{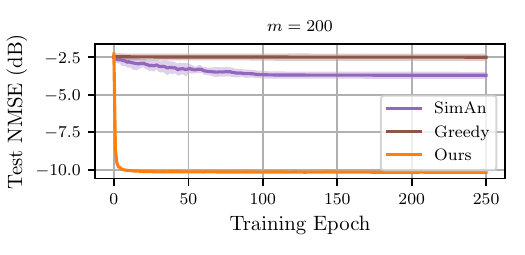}
\includegraphics[width=0.45\columnwidth]{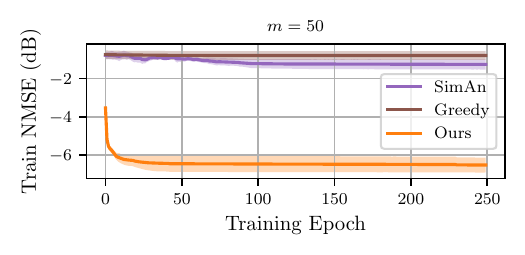}
\includegraphics[width=0.45\columnwidth]{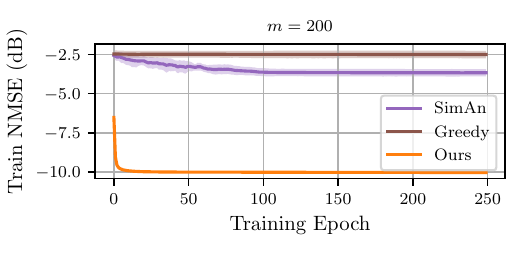}
\includegraphics[width=0.45\columnwidth]{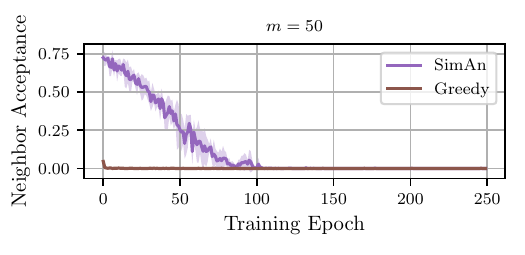}
\includegraphics[width=0.45\columnwidth]{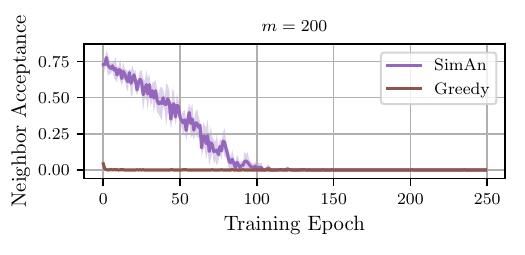}
\caption{Comparing the test NMSE (top), train NMSE (middle) and the average neighbor acceptance rate of the baseline algorithms throughout a training epoch (bottom) in the single pixel imaging setup from Section~\ref{experiments:section} with $m=50$ (left) and $m=200$ (right) measurements. The shaded area represents the standard deviation over 10 random seeds. This figure highlights the training stability of GLODISMO and shows that it is not overly dependent on the random seed. Furthermore, the bottom plots show that the baselines are reasonably tuned.}
\label{baseline:ablation}
\end{center}
\end{figure}

See Figure~\ref{baseline:ablation} for our supplemental results.

\section{Learning Fast-Transformable Structured Sparse Matrices}
\label{appendix:fasttransforms}
In many real-world applications, in particular when $m$ and $n$ are large, it is imperative that multiplication with $\Phi$ can be done via a fast transform algorithm in $O(n \log n)$ operations instead of a full matrix multiplication which would require $O(mn)$ operations. In this section we demonstrate how GLODISMO can be easily extended to learn discrete row-wise masked circulant matrices, which are a prime example of a class of matrices for which fast transforms exist.

As circulant matrices are diagonalized by the Discrete Fourier Transform (DFT), a row-wise masked circulant matrix $\Phi$ can be written as $\Phi = P_\Omega \mathcal{F}^{-1} \Lambda \mathcal{F}$ for a diagonal matrix $\Lambda$, DFT matrix $\mathcal{F}$, and a row-selection mask $P_\Omega$. This allows for multiplication in $O(n \log n)$ with $\Phi$ by using the only the DFT and point-wise vector multiplication. While it has been shown that certain random row-wise masked circulant matrices fulfill the restricted isometry property with high probability \cite{circulant}, GLODISMO provides an approach for incorporating structural constraints and training data priors into the measurement operator.
Our approach can be easily extended to work with any Toeplitz matrix, for which fast transforms based on the FFT also exist.

\begin{figure}
	\begin{center}
		\centerline{\includegraphics[width=0.8\columnwidth]{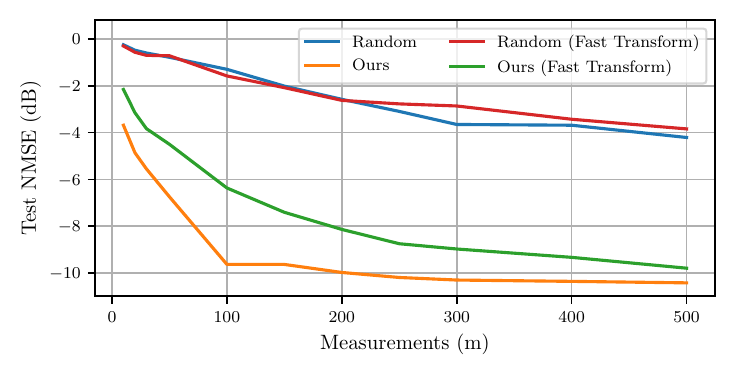}}
		\caption{Comparing the NMSE on the MNIST test set of fast-transformable single pixel imaging matrices with $d=32$ to regular single pixel imaging using IHT with $T=15$ iterations.}
		\label{fasttransform:varym}
	\end{center}
\end{figure}

\begin{figure}
\begin{center}
            \begin{subfigure}{.49\columnwidth}
		    \includegraphics[width=\columnwidth]{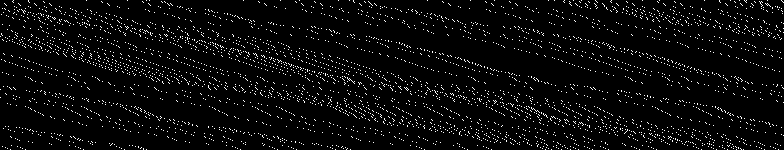}
		    \caption{Random, $m=150$\vspace{.5\baselineskip}}
            \end{subfigure}
            \begin{subfigure}{.49\columnwidth}
			\includegraphics[width=\columnwidth]{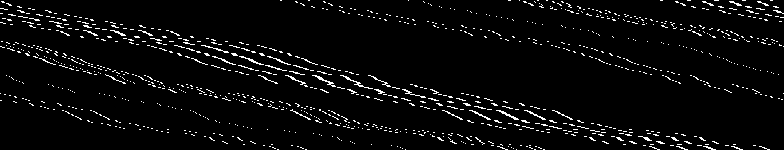}
			\caption{Learned, $m=150$\vspace{.5\baselineskip}}
            \end{subfigure}
            \begin{subfigure}{.49\columnwidth}
			\includegraphics[width=\columnwidth]{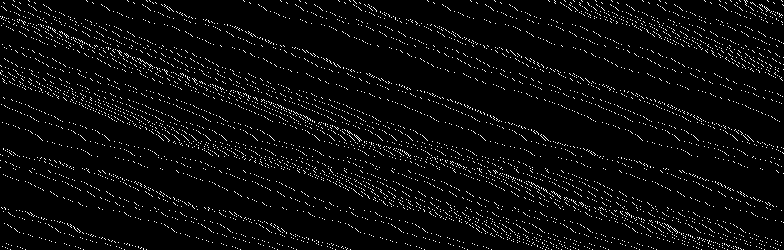}
			\caption{Random, $m=250$}
			\end{subfigure}
            \begin{subfigure}{.49\columnwidth}
			\includegraphics[width=\columnwidth]{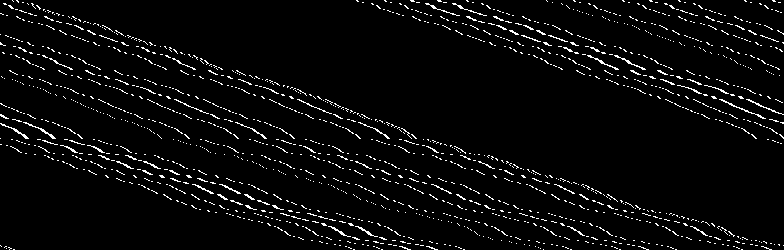}
			\caption{Learned, $m=250$}
			\end{subfigure}
\end{center}
\caption{Row-masked binary circulant single-pixel imaging measurement operators with $d=32$ ones per row. The learned fast-transformable operators clearly exhibit structure.}
\label{convolution:pics}
\end{figure}

In our setup, we use GLODISMO to learn binary row-wise circulant matrices with $d=31$ ones per row for single pixel imaging. We employ a Gumbel top-$d$ operation on a vector of length $n$, and use this vector to explicitly construct the circulant matrix. With a second Gumbel reparametrization, we learn the row-selection mask $P_\Omega$ that selects $m$ out of $n$ rows. Then, $\Phi$ is obtained by masking the circulant matrix with $P_\Omega$.
See Figures~\ref{fasttransform:varym} and~\ref{convolution:pics} for our results.


\section{Learning Super-Pixel Masks}
\label{appendix:superpixel}
When using single-pixel cameras for imaging at wavelengths which are in the same order of magnitude as the pixel-sizes, strong diffraction effects appear leading to poor image quality \cite{terahertzdiffraction}. It has been observed that using super-pixel masks, in which multiple adjacent pixels are always selected together, can mitigate this effect \cite{microscanning}. We demonstrate that it is possible to learn such super-pixel masks using a Gumbel reparametrization by using a simple modification of our method for learning a pixel mask.

The procedure for obtaining a super-pixel mask is as follows: for a fixed number of super-pixels $d$ with super-pixel height and width of $\Delta$, we follow the same procedure as when learning a pixel mask for determining the centers of the super-pixels. Each row of the matrix, corresponding to one mask, is then reshaped to image dimensions and the centers of the super-pixels are then convolved with a 2D convolution kernel of size $\Delta \times \Delta$ whose values are all ones. Then, the masks are reshaped back to rows of $\Phi$ and the maximum values of $\Phi$ are clipped to 1. Computing the gradients with regards to the learned Gumbel parametrization is then done via automatic differentiation.

\begin{figure}
	\begin{center}
		\centerline{\includegraphics[width=0.8\columnwidth]{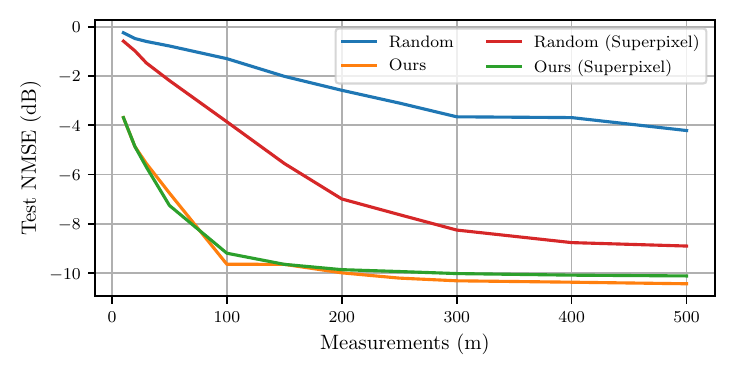}}
		\caption{Comparing the NMSE on the MNIST test set of super-pixel single pixel imaging matrices with $d=32$ to regular single pixel imaging using IHT with $T=15$ iterations.}
		\label{superpixel:varym}
	\end{center}
\end{figure}

\begin{figure}
	\begin{center}
		\begin{subfigure}{.95\columnwidth}
				\includegraphics[width=0.24\columnwidth]{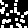}
				\includegraphics[width=0.24\columnwidth]{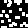}
				\includegraphics[width=0.24\columnwidth]{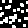}
				\includegraphics[width=0.24\columnwidth]{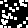}
				\caption{Random Masks\vspace{.5\baselineskip}}
        \end{subfigure}
		\begin{subfigure}{.95\columnwidth}
				\includegraphics[width=0.24\columnwidth]{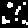}
				\includegraphics[width=0.24\columnwidth]{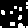}
				\includegraphics[width=0.24\columnwidth]{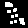}
				\includegraphics[width=0.24\columnwidth]{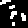}
				\caption{Learned Masks\vspace{.5\baselineskip}}
        \end{subfigure}
		\begin{subfigure}{0.95\columnwidth}
			\centerline{\includegraphics[width=.8\columnwidth]{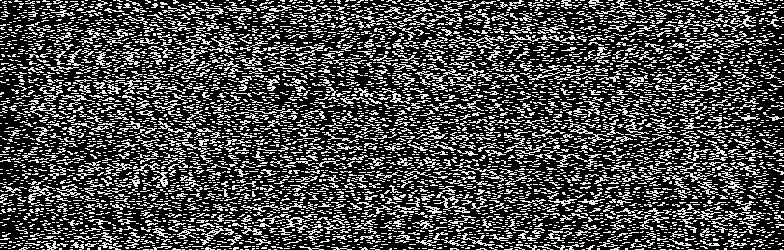}}
			\caption{Random $\Phi$\vspace{.5\baselineskip}}
        \end{subfigure}
		\begin{subfigure}{0.95\columnwidth}
			\centerline{\includegraphics[width=.8\columnwidth]{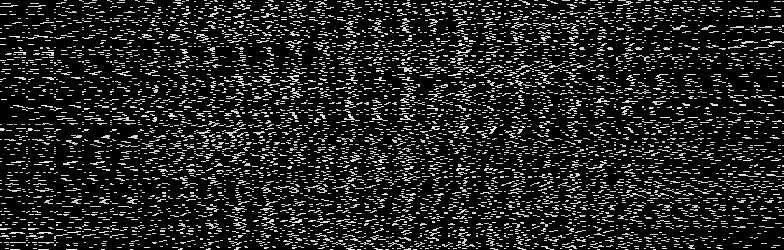}}
			\caption{Learned $\Phi$}
        \end{subfigure}
		\caption{Super-pixel masks for single pixel imaging with $d=32$ super-pixels per mask. The masks of the learned $\Phi$ clearly exhibit structure adapted to the MNIST dataset.}
		\label{superepixel:pixel_masks}
	\end{center}
\end{figure}

We employ this technique in the same MNIST single-pixel imaging setup as in Section \ref{singlepixel:section}. The results are shown in Figure~\ref{superpixel:varym}. Single-pixel imaging using super-pixel masks learned with GLODISMO outperforms using random super-pixel masks by a significant margin and match the performance of the learned pixel-wise masks. Figure \ref{superepixel:pixel_masks} shows how the learned masks adapt to the underlying dataset.

\section{Learning Pooling Matrices for Group Testing}
\label{appendix:grouptesting}

Consider an array of substances that are to be chemically tested for the possession or absence of a certain rare property. Examples are detecting disease in pandemic situations or detecting tainted products in chemical production chains. It is desirable to detect as many positives with the least amount of tests possible to economise on the tests as well as the personnel conducting the tests. The mathematical field of group testing is concerned with reducing the number of tests needed to faithfully discover all true positives by pooling the substances together. Here we consider a non-adaptive group testing approach in which all tests are conducted in parallel, and therefore the outcomes of the tests are mutually independent. Because usually only a small number of tests are positive, an array of samples to be tested can be modeled as a sparse vector. For $n$ samples (of which $s \ll n$ are positive), and $m$ tests, the $ij$-th component of the measurement matrix $\Phi$ for non-adaptive group testing then describes whether specimen $j$ is pooled into test $i$. As the tests are commonly standardized, it is sensible to model the number $d$ of specimen per test (i.e. the number of ones per row) to be fixed. The quantity to be measured in the individuals is commonly modeled to be non-negative, as chemical tests in general measure physical quantities. Therefore,  the inverse problem of identifying the infected individuals from the pooled tests can be solved using non-negative least absolute deviations (NNLAD) \cite{petercovid19}, that is
\begin{equation*}
    \min_{\hat{x} \geq 0}\|\Phi \hat{x} - y \|_1.
\end{equation*}
This convex optimization problem can be solved using a projected subgradient descent method as described in \cite{nnlad}, which can be unrolled and back-propagated through out of the box. The crucial question remains how to construct a $\Phi$ that can faithfully recover the positives from as few tests as possible. It is well-established that random matrices can be employed \cite{guruswami2009unbalanced}, when the conditions on $m,n$ and $s$ for compressed sensing are satisfied. Here, we follow the same problem size as considered in \cite{petercovid19}, which assumes $m=248, n=961$, and $d=31$. The measurement vector $y$ is contaminated by Gaussian noise equivalent to a signal-to-noise ratio of 40dB. 
In these experiments, the size of the support $s$ of the synthetic sparse vectors is fixed as $s=80$ (which is far above the value for which theoretical compressed sensing guarantees hold). More specifically, $s$ non-zero components are determined randomly without replacement. Mimicking viral load distributions, the non-zero components follow a heavy-tailed Beta distribution, where $x_i^{\text{nonzero}} \sim_{i.i.d} \text{Beta}(2, 8)$. We consider an individual to have tested positive if the value of the reconstruction is higher than 0.01 at a given index. The hyperparameters of the NNLAD algorithm were hand-tuned and determined to be $\sigma=0.1, \tau=0.6$.

\begin{figure}
\begin{center}
\includegraphics[width=0.45\columnwidth]{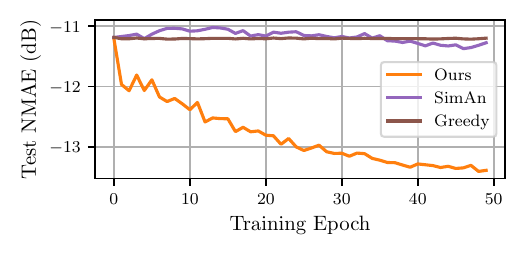}
\includegraphics[width=0.45\columnwidth]{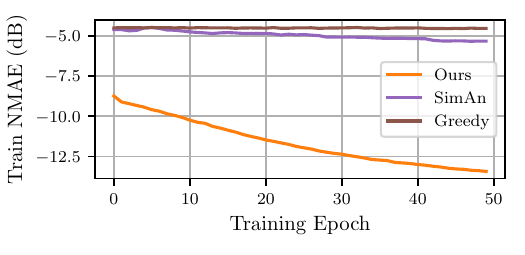}	\includegraphics[width=0.45\columnwidth]{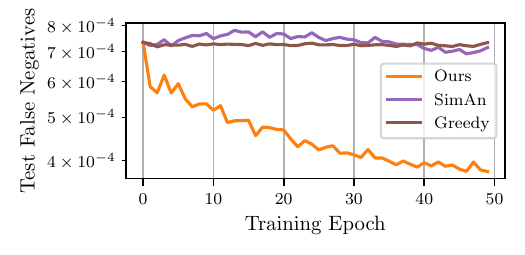}
\includegraphics[width=0.45\columnwidth]{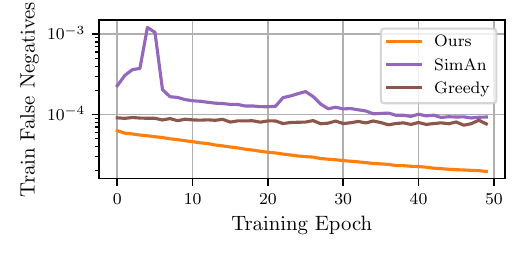}		\includegraphics[width=0.45\columnwidth]{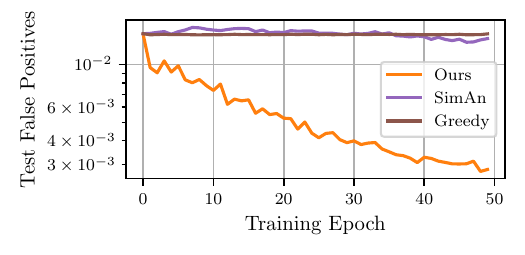}	
\includegraphics[width=0.45\columnwidth]{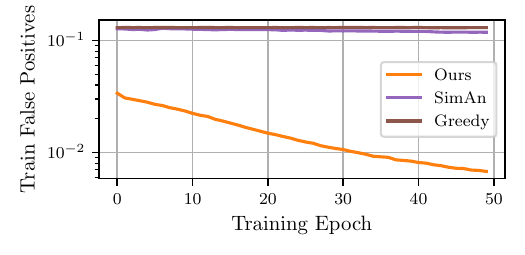}	
\caption{NMAE (top), false negatives (middle) and false positive (bottom) at training time with $T=200$ iterations (right) and during inference with $T=1000$ iterations (left) of the unrolled NNLAD algorithm. Despite the train/test mismatch, GLODISMO is able to learn a $\Phi$ that outperforms the baselines.}\label{pooling:synthetic}\end{center}
\end{figure}

In group testing, the time taken to conduct the tests is usually orders of magnitudes higher than the time required for solving the recovery problem algorithmically. In particular, the number of iterations taken until convergence is insignificant in group testing. At the same time, learning a pooling matrix with a large number of iterations is infeasible. Therefore, we learn $\Phi$ using projected subgradient descent for NNLAD \cite{nnlad} with $T=200$ iterations, and do inference with 1000 iterations, resulting in a mismatch between the train and test settings. To stabilize GLODISMO when training with 200 iterations, the loss is averaged over the reconstruction after every iteration and not just the final estimate.
Our results can be found in Figure~\ref{pooling:synthetic}.

\bibliographystyle{IEEEtran}
\bibliography{references}

\end{document}